# Irrespective Priority-Based Regular Properties of High-Intensity Virtual Environments

Kirill A. Sorudeykin, *Member, IEEE*

*Abstract* — We have a lot of relation to the encoding and the Theory of Information, when considering thinking. This is a natural process and, at once, the complex thing we investigate. This always was a challenge – to understand how our mind works, and we are trying to find some universal models for this. A lot of ways have been considered so far, but we are looking for Something, we seek for approaches. And the goal is to find a consistent, noncontradictory view, which should at once be enough flexible in any dimensions to allow to represent various kinds of processes and environments, matters of different nature and diverse objects by the singular apparatus. Developing of such a model is the destination of this article.

*Keywords* — Space of Thinking, Theory of Interactions, Principles of Mind, Mechanisms of Reasoning, Natural Intelligence, Algorithm, Path of Execution, Architecture.

## I. Introduction

DOES the modern technology account the possibilities which our mind grants? On what way the technology development is directed? How to describe what is lying under the natural processes we observe, what is common and what is different between them? What is the basis? And, which is more important, what is the matter we are dealing with, conducting programs, making devices, trying to control?….

The world we see is unbounded. But we observe boundaries. Boundaries of time, boundaries of movement, boundaries of resources… They appear eventually, appear as a result of any other interactions, as the properties of media… We cannot even say when does they appear, but we are used up to treat them as the usual properties of reality. As the fact, as we observe them. But they are not.

In the paper we consider the properties of reality which give us the boundaries, which, in turn, define the real interaction of things, resulting in what we treat as *events*. In result of this the notions appear and a coordinate system can be applied to the space, forming what we call *topology*. Then to this the *inertia* property being added which lead to the real *tunneling*, interaction and *concurrency* appearance. Then we receive some kind of tight communications with density, in which we can find *finite paths* of execution.

This tightness lies very close to the *optimal architectural design* and *classification*. Naturally, some parts of

The author Kirill A. Sorudeykin is with the Kharkov National University of Radio Electronics, Ukraine and Tallinn University of Technology, Estonia. At once he is the Chair of Relevance Research & Development Corporation. E-mail: Kirill.Sorudeykin@relvecorp.com, Kirill.A.Sorudeykin@ieee.org

systems can be *parallel* in relation to each other. And can require *scheduling*. The parts can be in concurrency. And the mechanisms of their interactions, as well as the structure of their appearance represent what we call nature. These mechanisms are irrelated in basis to a field of use, i.e. they are universal and deserve detailed studying.

We would like to consider here how the principles of thinking, which are by the way the display of some properties of reality, can be translated to a language of reality, clearly translated. To see this without the relation to something biological or social, but grounding on something more profound. Deeper than the things we got accustomed to in our everyday's life. What stays behind the successions and simultaneousness, behind specifics and separation, behind objects, locations and order.

By this we could explain what the terms *understanding* and *realization (awareness)* mean, which are related to the *decisions making.* Then we will be able to express what *meaning* and *thinking* themselves are.

In the next section we will describe the preconditions for the research from the analytical point of view on processing and computational capability.

## II. Problem Description

> *"Action expresses priorities."*
> Mahatma Ghandi

Let's imagine that we are looking on the very basis of computational processes. Then we should see some departing point of a task and some processing unit (as of today's architecture of computations). This unit will make something like bottleneck (Fig. 1) which will let the task or data pass through it according to the capacity of this unit. Anotherwords this unit will limit the computational abilities of the system in some way, letting the process flow serially:

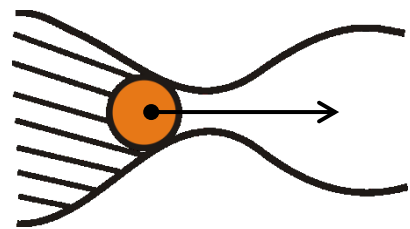

Fig. 1. A bottleneck.

Something wider should be expressed by means of something narrower. The same situation in various forms we observe in nature and to a large extent it determines the look of the real world and all the possible existence we can observe. Anotherwords, it determines the reality. The matter is that if we talk about *interactions* in some space, we need to consider *concurrent* objects with defined shapes and boundaries in relation to surrounding environment. And this imposes limitations on the rules according to which this space can be (or needs to be) built. But let's look onto this property more closely.

If we have a certain idea, we need to explain it in some way to let it be implemented. Even if we are going to implement this idea by ourselves, not telling about it to anybody, we still need to *express* it in a certain language – the language of the *media* in which you want to shape it. For example, sculptor needs to implement a new sculpture in marble, - so they need to express the idea of the sculpture on the language of marble, i.e. on the language of stone, which has its own specific set of expressive means, as well as the technique of treatment, i.e. matching a shape of a stone to achieve the desired artistic impression.

### III. LIMITATIONS OF PROCESSES

The idea, as we already specified above, will be limited in order to be expressed by the language we use in reality:

$$Idea \cap [Lang] = Expr$$

Where an *Idea* have the following property:

$$Idea = \{P, E\}$$

$P$ – point, $E$ – e-neighborhood

As we see, this is the same situation we have outlined earlier – some limiting factor that prevents expressing of something one-to-one. And a distortion (change) of initial idea can appear in result of inability of expressive means to reproduce the idea absolutely identically. Marble possibly will not allow to carry out some things. For some other things this material can even not be proper or at least will not be the best solution. For example, this is a brittle material, that's why some subtle elements cannot be at once strong. And the implementation ability of especially delicate things will depend on a grain size of the material. So the author should care about the combination of design and functionality. At once this is not so cheap material, that's why the product should correspond.

In modern computer systems such term as *bus* can be met by which the communication channel between some functional blocks of the system can be denoted. As well as the block themselves, busses as communication channels also limit the expression possibilities of data to their widths. They can transfer just "portions" of data in a time, not the whole image, which became one of the main limitations of current processors. And if we analyze further we could see that any of processing unit, whether it is a processor, or person, or even a physical object, like tree or stone, should be defined in order to exist. And being existent, it became inertial to change, because the parts are inertial to change, as well as all the surrounding objects.

We could call the interface of the module a *perception window*, because this term can characterize not only processors or functional elements, but also any abstract object. And we will take it by this beyond the scope of specifics. This term was analyzed in the paper [1]. So, considering more clearly we are obtaining such an expression:

$$D \cap B = q$$

$D$ – Data
$B$ – Bus (perception window)
$q$ – Quant

As we see, bus size limits the transition capabilities of data from the departure to the goal point. As a result we are obtaining the series of quants, representing a data or a process. This is the basis for time slicing technique (approach), scheduling, preemption and queuing, widely used in today's technology [2], [3].

And quants together represent a series, i.e. a set of quants, which needs to be processed:

$$D(B) = Q$$

$$Q = \{q_i\}, i = \overline{1..n}$$
$$P(q_i) = t_i;$$
$$P(Q) = T$$

$P$ – function of power
$t$ – time for a singular quant
$T$ – the whole time of the processing

Anotherwords, as a result of quantification we obtain a time of the serial processing. Limitation of a processor, as a functional block, is limited capability, and therefore time necessity for serial execution. Then we have:

$$D \rightarrow \bigcup_{i=1}^{n} q_i = D'$$

$D'$ – new representation of data

i.e. data representation changes (being transformed) after passing through the bus, as we noted earlier, and we are obtaining a synthetical representation of our departing point, which as usual will not carry all the specifics of the initial data (initial image of a task). In general case:

$$D \neq D'$$

And what is the difference?

$$D \oplus D' = I_A$$

$I_A$ – Influence of the architecture

Here we see the difference as a character of the architecture which in turn represents a structure of threads in a system, according to which quantification is taking place and influences on data. Shortly saying, this indicator is the image of data change in result of the architecture's influence. The representation of data changes, i.e. in general case we cannot display the image from one world (or *virtual space*) one-for-one to another world. That's why something being loosed and we need to recoup this with the additional description on the language of target space in order to ensure that the resulting image will be enough full to transfer a *main sense* of the initial image, which means additional space and time. But even in that case it will not be full absolutely, because of incompatibility (even if a small difference) of two spaces (languages).

The architecture itself is a layout of the subject field, possible ways of designing something, the structure of the media. It is much wider space than local solution and the relation of it with the value $I_A$ that characterizes the causes of data change (distortion, transformation) is as following:

$$I_A = (A_S \subseteq A)^{-1}$$
$$A_S = \{s: D \mapsto A\}$$
$$D \overrightarrow{\cap} A = [D(A)] = I_A$$

$A_S$ – specified architecture, subset of $A$
$\overrightarrow{\cap}$ – non-commutative (one-side) intersection

I.e. $I_A$ is the inversion of a specified architecture. It comes as a result of finding a specific way for implementing D in a desired space, taking all the peculiarities of $A$, which D faces. And, if representing the experience' collection procedure we could express the following:

$$D = \{d_i\} \text{ – set of local tasks}$$
$$\forall d_i: A = A \cap d_i \Rightarrow A = \bigcap\{d_i\}$$

Which is the individual cyberspace (individual architecture) for the specific set of tasks (data). The problem is that we could specify the initial set $D$ only on the stage of the development of individual solution, making it as much universal and general-purpose, as desired. But when other systems rely on ours one, we cannot change (adjust) it so easily, i.e. we cannot change its principal behavior quickly enough without attracting the change of the whole system. That's why the principal architecture should be designed most carefully, satisfying the condition (requirement) of absence (limit) or minimally-possible (on practice) amount of redundancy, which is at once economic, ordered, lying closer to the original image, so transfers the most of its sense, and has a lot of other benefits, related to optimality.

IV. ARCHITECTURE-DIRECTED OPTIMIZATION

The optimal representation of space maximizes at once all the considered characteristics independently of each other. This situation is called a Pareto optimum. And we could find the *fixed point* (tension point), i.e. a Nash solution, where all the characteristics being maximized taken together as a system. In the papers [4]-[6] we already considered some topics of this research, for example fractal sets of limitations, let's continue this analysis now.

If we need optimality, we need anotherwords an architecture that has the minimal impact to the data, therefore to the original task (image), i.e. the specified architecture:

$$\Rightarrow \quad I_A \to min \quad \Leftrightarrow \quad A \mapsto O$$
$$O \sim D \text{ ($O$ is related to $D$)}; A \neq O$$

$O$ – Original view of a task

Here an original view $O$ of a task (idea $I$) corresponds to the initial form of data $D$, and as we see, the architecture should strive to this original form to be based exactly on that idea and coincide as a space with it. In this case the space of architecture should repeat the space of experience, where the idea has arisen. And then we have:

$$A \overrightarrow{\oplus} O = L$$

$L$ – limit of space

I.e. architecture becomes a limiting factor in relation to the original view on the task. Difference between two spaces of views displays this. Here we use our new notation – the *implementation* operator (the arrow, directed contrariwise to the operand we continue to talking about and which will be implemented in a result) with the *operation* (specified under the arrow) by means of which the implementation will be performed. So, what we need to do to make a process quicker?

$$\text{For } D \overrightarrow{\cap} B = Q; Q = \{q_i\}; i = \overline{1..n}; n \sim T$$
$$\text{If } T \to min \text{ we need } n \to min \text{ [and } t \to min]$$

Because of $D = const$ in our case, as original form of data, our image, we need to change $B$, the specificity of the architecture. Doing this we should obtain such a *native* decomposition of data (i.e. derivable from it), not to break its own compositional structure. We came from the single simple statement of quantification to the system of statements which characterizes this in a more deep and flexible way. In this system quants being determined by the data in order to transfer natural blocks of data in a time, not separating it artificially. Of course, component structure of data can get some influence of architectural specifics, but should not have it too much. And, as we see, senses of bus and quantization change. Here is our new structure:

$$D \overrightarrow{\cap} B = Q \Rrightarrow \begin{cases} D' = \{d_i\}; \ B' = \{b_j\}; \\ D \mapsto (D' \dashv D); i = \overline{1..n}; \\ (B \sim D') \lessdot B'; j = \overline{1..m}; \\ P(B) \to [f(b_j) \leq |b_j|]; \\ \forall b_j \geq [\sigma_{d_i \sim b_j}(D') \lessdot D']; \\ D - Pareto\ optimal. \end{cases}$$

$D$ – Data,
$B$ – "Bus"
$P(B)$ – power of $B$ (capacity), $|B|$
$... \lessdot D'$ – is an aggregation part of $D'$

The architecture, as we already noted, should be derivable from the informational model of the subject domain, so it should be flexible. But by what means we can achieve this? Modern architectures have other specifics. But we need something which should be able to evolve. In response to this hereby we are stating that the *main* direction in our research is the *flexibility of architectures* (processor and computer architectures, computational architectures, software architectures). This should involve the very computational disciplines and approach to computations itself. Then it should reflect on the means we use to compute. They should show the ability to structure the information in a specific way. Computational discipline should not be determined by the actions related just to specific objects, but it should be determined by the data and the basic set of rules of evolving of the space of knowledge. I.e. computational process itself should be generic and emerging in principle and look more like the spatial interactivity than streaming. And the architecture should also have the features of such interactivity, whether we are talking about hardware or software architectures, which represent just different levels of logic.

## V. Comprehension-based Constructions

In this section we will give the description of the architectures with the new properties we are developing. Let's consider the set of solutions *S*. Each individual viewpoint has several such solutions in a singular viewspace, and each of those solutions consists of some components, which reflex its inner content:

$$S = \{s_k \mid s_k = \{elt_m \mid elt_m \in SubjFld\}\}; \; k, m = \overline{1..r}$$

The goal of solutions is to identify more and less stable (static) elements in the environment to build the right strategy in accordance to the circumstance, i.e. to determine the rules. Corresponding to the level of statics (inertia) we can determine the value (weight) of the elements. We also can see the inclusion relation between them. For example, let's consider two objects *A* and *B* belonging to different layers of inertia 1 and 2: $A^1 \sim B^2$. Each of these objects has {weight_of_move and time_of_move} as the properties: $A^1 = \{m, t^1\}; B^2 = \{n, t^2\}$. If we try to apply the value of $B^2$ to the value of $A^1$, the following relation can be taken: $[\, n \to (\alpha \cdot n \approx m); t^2 \to (\beta \cdot t^2 \ll t^1) \,] \Rightarrow B^2 < A^1$

Subject field is the source of knowledge we are using to build a model. And the model *M* should contain all the solutions, common part of which should be joined and the rest added uniquely:

$$M = \bigcap_k s_k + \bigoplus_p s_p$$

This expression displays the structure of packed set of solutions, the space of experience. Each space of experience should contain the intersection of factors, which will point to a specific notion. And this intersection has the specific structure, i.e. some form of organization, which reflects certain specificities of organized spaces (Fig. 2). Works [6], [7] show different algorithms on graphs and they are good examples of approaches to store the experience and make the classification of categories in such structures.

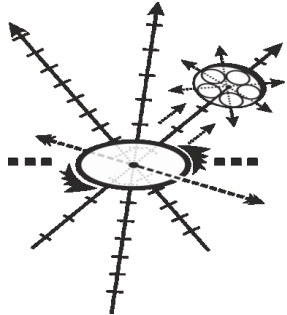

Fig. 2. An intersection of the categorical factors.

References [8]–[12] show very interesting topics on the study of thinking and mind. Such works form today's views on the nature of intelligence and possibilities of our mind. But they contain something more – a hidden idea of future theories, invisible message, a hint onto something still unresolved, but what feeling of authors tried to express. We here are analyzing those topics to give the further development of their ideas and throw the light upon the thrilling questions of science of today.

## VI. Conclusion

This paper continues the research, performed in [1], [4]–[6] and here we have made a step forward in formalization of irrespective properties of observable and imaginable spaces. This research can help us to understand more clearly what mind is, and we will not be needed anymore to address thinking just to the physical processes in specific tissues or organisms, or even to some specific physical objects, and will be able to look beyond this, to the elementary properties of nature, which form all the variety of other possible displays of reality and virtual spaces, which can be both observable, existent, or imaginable.

In further research we will continue to specify the spatial properties of reality and will continue to formalize what we found. The goal will be to join what we already have, to extend the basic notions and terms, to cover more advanced topics, to finish the basic variant of a full model and obtain practical results in decisions making, planning, analysis, automated design, effective reasoning, optimization or using this theory in education.

## References

[1] Kirill Sorudeykin, "Levels of Ordering in Software Design and Thinking", *Proceedings of IV-th all-Ukraining conference "Intellectual Computer Systems And Networks"*, Krivoy Rog, Ukraine, 2011, pp. 178-181.
[2] Ivo Adan and Jacques Resing, "Queueing theory". *Dep.of Math. and Comp. Sci., Eindhoven University of Technology, Netherlands*, February 28, 2002, 180 p
[3] Yaashuwanth .C & R. Ramesh , "Intelligent time slice for round robin in real time operating systems", *Int. Jour. of Res. and Rev. in Appl. Sci. (IJRRAS)*, vol. 2, no. 2, February 2010 pp. 126-131
[4] Kirill A. Sorudeykin "A model of Spatial Thinking for Computational Intelligence", *9th IEEE East-West Design & Test Symposium, (EWDTS'11)*, Sevastopol, Ukraine, *2011*pp. 311-318
[5] Kirill Sorudeykin. "An Operational Analysis and the Degree of Inertia in Thinking Process Modeling". *Proceedings of VI-th international conference "Science And Social Problems of Society: Informatization and Informational Technologies"*, KhNURE, 2011, pp. 335-336
[6] Kirill Sorudeykin, Valentina Andreeva. "A Research of Heuristic Optimization Approaches to the Test Set Compaction Procedure Based On a Decomposition Tree for Combinational Circuits", *10th IEEE East-West Design & Test Symposium, (EWDTS'12)* Kharkov, Ukraine, *2012,* pp. 382-387
[7] Charles E. Leiserson, Ronald L. Rivest, Clifford Stein "Introduction to algorithms, second edition", *The MIT Press*, 2002, 1290p
[8] George A. Miller*,* "The Magical Number Seven, Plus or Minus Two. Some Limits on Our Capacity for Processing Information"*, Psychological Review*, vol. 101, no. 2, 343-352
[9] Georgi Lozanov, "Suggestopaedia – desuggestive teaching. Communicative method on the level of the hidden reserves of the human mind"*, Dr. Georgi Lozanov, International Centre for Desuggestology*, Vienna, Austria, 2005, 140p
[10] Charles S. Peirce, "On a New List of Categories", *Proc. of the American Academy of Arts and Sciences vol.* 7, 1868, pp. 287–298
[11] C. S. Peirce, "Reasoning and the Logic of Things", *The Cambridge Conference Lectures of 1898, Harvard*, 1992, 312 p.
[12] Jacques Hadamard, "An essay on the psychology of invention in the mathematical field", *Courier Dover Publications*, 1954 – 145 p.